\documentclass[journal]{IEEEtran}

\usepackage{caption2}
\usepackage{color}

\ifCLASSINFOpdf
   \usepackage[pdftex]{graphicx}
   \DeclareGraphicsExtensions{.pdf,.jpeg,.png}
\else
\fi
\usepackage{multirow}

\usepackage[cmex10]{amsmath}

\usepackage[ruled,norelsize]{algorithm2e}
\SetKwRepeat{Do}{do}{while}%

\makeatletter
\newcommand{\removelatexerror}{\let\@latex@error\@gobble}
\makeatother

\usepackage{array}

\ifCLASSOPTIONcompsoc
  \usepackage[caption=false,font=normalsize,labelfont=sf,textfont=sf]{subfig}
\else
  \usepackage[caption=false,font=footnotesize]{subfig}
\fi

\graphicspath{{images/}}

\begin{document}
%
\title{Hierarchical Spatial Sum-Product Networks for Action Recognition in Still Images}
%
%
%

\author{Jinghua~Wang, Gang~Wang*,~\IEEEmembership{Member,~IEEE,}
	\thanks{*Corresponding author

		Jinghua Wang and Gang Wang are with the School of Electrical and Electronic Engineering, Nanyang Technological University, Singapore (email: jinghuawng@gmail.com; wanggang@ntu.edu.sg)
		
		Copyright (c) 2016 IEEE. Personal use of this material is permitted.
		However, permission to use this material for any other purposes must be obtained from the IEEE by sending an email to pubs-permissions@ieee.org.
	}
}

%
%

\markboth{IEEE TRANSACTIONS ON CIRCUITS AND SYSTEMS FOR VIDEO TECHNOLOGY,~Vol.~xx, No.~xx}%
{Shell \MakeLowercase{\textit{et al.}}: Bare Demo of IEEEtran.cls for IEEE Journals}

%



\maketitle

\begin{abstract}
Recognizing actions from still images is popularly studied recently. In this paper, we model an action class as a flexible number of spatial configurations of body parts by proposing a new spatial SPN (Sum-Product Networks). First, we discover a set of parts in image collections via unsupervised learning. Then, our new spatial SPN is applied to model the spatial relationship and also the high-order correlations of parts. To learn robust networks, we further develop a hierarchical spatial SPN method, which models pairwise spatial relationship between parts inside sub-images and models the correlation of sub-images via extra layers of SPN. Our method is shown to be effective on two benchmark datasets.

\end{abstract}


%
\IEEEpeerreviewmaketitle

\section{Introduction}
%
%
%
%
%
%

\begin{figure}[h]
	\begin{center}
		\includegraphics[width=0.9\linewidth]{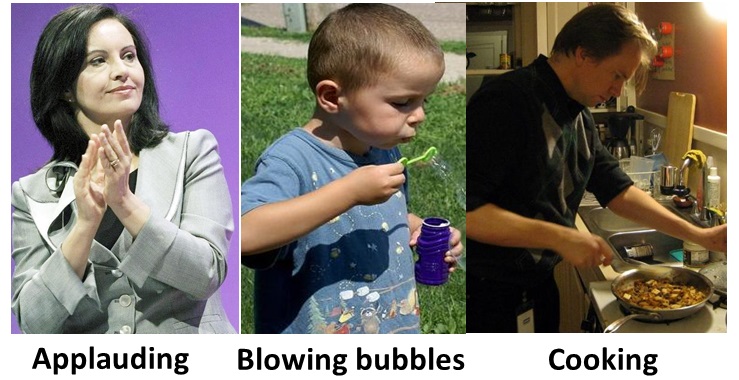}
	\end{center}
	\caption{Example images from Stanford40 dataset \cite{Yao11humanaction__Stanford40}. We humans can easily recognize the action class of these three images (applauding, blowing bubbles, and cooking).}
	\label{fig:introductionExample}
\end{figure}

\IEEEPARstart{A}{ction} recognition from videos has been an active research topic in computer vision for more than two decades \cite{computers2020088,TCSCVT_1,TCSVT_2,TCSVT_3,TCSVT_4,reviewer1_6918650,reviewer1_Baccouche:2011:SDL:2177908.2177914,reviewer1_Dobhal2015178}. However, video is not essential in action recognition. Based on the three images in Fig \ref{fig:introductionExample}, we human beings can easily identify the action classes, i.e. \textit{applauding, blowing bubbles}, and \textit{cooking}. This observation motivates the computer vision community to develop techniques for action recognition from still images, which has many potential applications in image annotation, image retrieval, and video-based action recognition \cite{StillImageSurvey}.

To recognize action classes from still images accurately, researchers tend to integrate it with the task of pose estimation \cite{YangWMCVPR10,Yao_modeling_mutual_2010_CVPR}.
In the integrated framework, these two tasks can help each other. 
However, the performance of action recognition heavily relies on the pose estimation result. The failure of pose estimation can significantly reduce the action recognition accuracy. Other interesting methods recognize object-associated human actions by modeling the interactions between humans and contextual objects \cite{Delaitre11learningperson-object,GroupletYaoBangpengFeifeiLI}.
It has been shown that action recognition can also achieve high accuracy holistically, which can be applied in more general cases where the information of pose and associated objects is not available \cite{sharma:CVPR2012_Discriminative,Delaitre10recognizinghuman,sharma:CVPR2013}. 

Some other works \cite{Yao11humanaction__Stanford40,MajiActionCVPR11,sharma:CVPR2013,BourdevMalikICCV09,Yao_modeling_mutual_2010_CVPR} have shown that representing action classes by a set of body parts can overcome the limitation of the highly structured models.
Inspired by the success of part-based methods, we model an action class as a flexible number of spatial configurations of parts.
For image representation, we propose an unsupervised method to learn a set parts. Different from previous works that learn the parts based on low-level features, the proposed method discovers parts through deep feature clustering and CNN model fine-tunning. 
In the proposed method, deep feature clustering and CNN model fine-tunning can boost the performance of each other. This method can be directly applied to other tasks for unsupervised visual pattern discovery.

With a set of parts, an image can be represented by an activation vector of these parts, as well as their locations. The part activation vector is useful for action recognition, as it reveals which parts occur in the image. Informative parts indicate the potential class label of an image. For example, the parts that represent the appearance of a bike are expected to occur in an image from the action class of \textit{riding a bike}.
However, the same set of parts may occur in two different classes. It is the layout of parts that discriminates one from the other.   
Thus, the spatial relationships of these parts are also critical.
For example, the main difference between the action classes of \textit{riding a bike} and \textit{fixing a bike} is the spatial relationship between a human and a bike.


In order to incorporate the spatial relationship between parts in action classification, Desai et al. \cite{desai10_action} propose to mathematically model the locations of the parts. By implicitly assuming the spatial relationship between parts are fixed in the images of the same class, it cannot deal with deformable part pairs.
Differently, in this work, we propose spatial SPN (Sum-Product Networks) to capture the spatial relationships as well as high-order correlations of the parts.
SPN is first proposed in \cite{poon2011SPNIntroduce} to model the joint probability of variables in a hierarchical manner. In naive SPN, the spatial relationships between its inputs are completely ignored. To capture such important information in action recognition, we introduce four types of indicator nodes (\textit{right, left, above, and below}) for the product nodes which are the immediate parents of two part nodes. These indicator nodes encode the different spatial relationships between a pair of parts. Thus, the spatial SPN can deal with deformable spatial relationship between part pairs.

\newcommand{\slfrac}[2]{\left.#1\middle/#2\right.}

It is preferable to model local instead of long-range spatial relationship between parts in the task of action recognition due to two reasons. 
Firstly, local spatial relationship is more stable. The local relationship \textit{a knife above vegetables} always occurs in images from the action class of \textit{cutting vegetables}. Long range spatial relationship between parts, which can vary significantly, may not carry discriminative information. An example is shown in Fig \ref{fig:reading_example}. Humans can have different poses in images from the action class of \textit{reading}. 
Secondly, it is computationally expensive to model the spatial relationship of every pair of parts.
With $ N $ parts, there are $\slfrac{N(N-1)}{2} $ possible part pairs. Instead of modeling the spatial relationship of all part pairs, we hierarchically partition an image into sub-images and only consider the part pairs that co-occur in the same sub-image. In this way, we drop the long-range spatial relationship between the part pairs, and significantly simplify the structure of SPN.

We model the correlations of the sub-images using the top layers of the spatial SPN. To achieve this, we propose a method for SPN structure learning based on image partitioning.
In the proposed method, we hierarchically partition an image into sub-images (bottom layers model spatial relationship locally between parts inside sub-images). Among a large number of possible partitions, we only encode the discriminant ones in the spatial SPN for efficient learning. In the hierarchical spatial SPN, a product node is associated with a partition method. As the parent of several product nodes (associating with different partitions), a sum node models the combination of different partition methods, which collaborate to improve the discriminant ability.

In short, our main contributions are as follows:
1) we propose a new representation for action class in still images based on SPN;
2) we propose spatial SPN, a new structure of SPN, to model not only the high-order correlation, but also the spatial relationship between its leaf nodes;
3) we propose a new method to learn the structure of SPN based on image partition.
We test our method on two datasets (Willow 7 action and Stanford 40 action). The experiment results show the effectiveness of our method.

The remaining part of this paper is organized as follows. Section \ref{Sec:Related_work} describes related work. Section \ref{Sec:Approach} introduces the proposed method for part discovery and spatial SPN structure learning. Section \ref{Sec:experi} shows the experiments. Section \ref{Sec:Conclusion} concludes this paper.

\section{Related Work}
\label{Sec:Related_work}

To recognize actions from images, researchers propose methods to learn discriminative representations for humans under different poses. Ikizler-Cinbis et al. \cite{ikizlercinbisICCV2009} learn HOG-feature based representations for different action classes based on the images collected from Web. Thurau and Halvac \cite{Thurau-HlavacPosePrimitivesCVPR2008} train a set of pose primitives by non-negative matrix decomposition of HOG-descriptor and represent images using these pose primitives. Wang et al. \cite{DBLP:conf/cvpr/WangJDLM06} propose a technique for deformable matching of edges from a pair of images. These methods \cite{ikizlercinbisICCV2009,Thurau-HlavacPosePrimitivesCVPR2008,DBLP:conf/cvpr/WangJDLM06} extract features from the whole image and obtain a global template. However, these global templates are not effective for the action recognition due to the significant pose variations in images.

It has been shown that part-based methods are more robust than global-based methods against pose variations \cite{YangWMCVPR10,FelzenszwalbMR_CVPR_2008}. Bourdev and Malik \cite{BourdevMalikICCV09} introduce the idea of poselet for robust person detection.          
Later, Bourdev \cite{BourdevPoseletsECCV10} et al. propose to learn poselet based on 2D keypoints and take the spatial relationship of these poselets into consideration. Yang et al. \cite{YangWMCVPR10} integrate pose estimation and action recognition in a single framework. In this work \cite{YangWMCVPR10}, a poselet represents a set of patches not only similar in pose configuration, but also belonging to the same action class. To recognize human and object interactions, Yao and Fei-Fei \cite{GroupletYaoBangpengFeifeiLI} propose `grouplet' to capture the structure information of an image via an AND/OR graph.  Yao et al. \cite{Yao11humanaction__Stanford40} incorporate attributes that describe the properties of human action into part-based representation. For action prediction, an image is sparsely represented by a set of action bases. To recognize human-object interactions, Desai et al. \cite{desai10_action} represent an image by a set of overlapping patches at various locations with their HOG features. To define a contextual model, the patch features are linearly combined with the configuration structure. Maji et al. \cite{MajiActionCVPR11} propose a dataset with 3D pose annotations and represent an image by the pose activation vector. Sharam et al. \cite{sharma:CVPR2013} propose a SVM-like model to capture the spatial relationship between the parts for action recognition and attribute learning. However, it makes the model very complicated to linearly combine the HOG features with the locations of the detected templates.

SPN (Sum-Product Networks) \cite{poon2011SPNIntroduce} is a newly proposed deep structure that can capture the high-order correlations between its leaf nodes. A number of papers later investigate this structure theoretically \cite{Delalleau11shallowvs,DirectIndirectSPNicml2014c1_rooshenas14}. In addition, SPN is proved to be successful in several tasks of computer vision, including image classification \cite{Discriminative_Learning_SPNNIPS2012_4516}, facial attribute analysis \cite{SPNWangXiaogangFacial}, and action recognition in videos \cite{SPN_video_action}. 
In this paper, we characterize an action class by several configurations of body parts, and represent it using SPN.
Amer and Todorovic \cite{SPN_video_action} propose to learn a Bag-of-Words representation for a video and model the deep correlations of parts using SPN with a stochastic structure. However, the spatial relationships of the parts are not taken into consideration. 
For the first time, in this work, we propose spatial SPN to explicitly model the spatial relationship of parts for robust action recognition.

\section{Approach}
\label{Sec:Approach}

We propose to recognize actions based on parts, which are discovered in images by adapting CNN model in an unsupervised manner. The parts can be noisy, hence we propose to use SPN to model the relationship between parts robustly.
SPN \cite{poon2011SPNIntroduce} has been shown to be very effective for representing high-order correlations between variables. However, traditional SPN cannot represent the spatial relationship of parts, which is very critical for our action recognition problem. In this work, we propose a new hierarchical spatial SPN to model both spatial relationship and high-order correlations of parts for improved recognition.

\subsection{Part Learning}
\label{section_part_learning}

Part-based methods \cite{liuting} are popularly used for action recognition and achieve high accuracy \cite{YangWMCVPR10,MajiActionCVPR11,YiYangCVPR2011:APE:2191740.2192012,Yao11humanaction__Stanford40,sharma:CVPR2013,Shahroudy2015Multimodal}.
They are proved to be more robust than global-based methods against pose variations \cite{YangWMCVPR10,FelzenszwalbMR_CVPR_2008}.

In this work, we consider a part as a visual pattern that occurs in many images. It is a difficult task to identify parts, due to the missing information of: 1) the reference for a part; 2) whether a part occurs in an image or not; and 3) the location of a part (if it occurs). To overcome these difficulties, we develop an unsupervised learning method to discover parts in image collections.

Deep learning has attracted wide attention due to its great success in several tasks \cite{Zuo2015Learning,Shuai2016Scene}. Here, we aim to learn a CNN model which can predict the part label of an input image patch.
To obtain such a model, we start with the CNN model pre-trained on imageNet \cite{CNN__NIPS2012_4824,jia2014caffe}. It is proved that CNN can achieve higher image classification accuracy than shallow models \cite{CNN__NIPS2012_4824}. This indicates CNN-based deep features are more powerful in visual pattern representation.  

We densely sample $ n_p $ patches from each image for part discovery. We conduct the following three steps iteratively: deep feature extraction, unsupervised clustering, and fine-tuning. 
Firstly, we extract deep features from the fully-connected layer of CNN \cite{CNN__NIPS2012_4824,jia2014caffe}. Secondly, to obtain a tentative reference for each part, we conduct unsupervised clustering and obtain $ n_c $ clusters ($ n_c $ decreases in each iteration). Each cluster is considered as a tentative part. We take the center point of a cluster as the reference of this part. Thirdly, we fine-tune the CNN model with the cluster labels. In this way, the CNN model will be more effective in capturing the visual patterns of our data. The deep features extracted in the next iteration can fit our task better.

\begin{figure}[t]
	\begin{center}
		\includegraphics[width=0.85\linewidth]{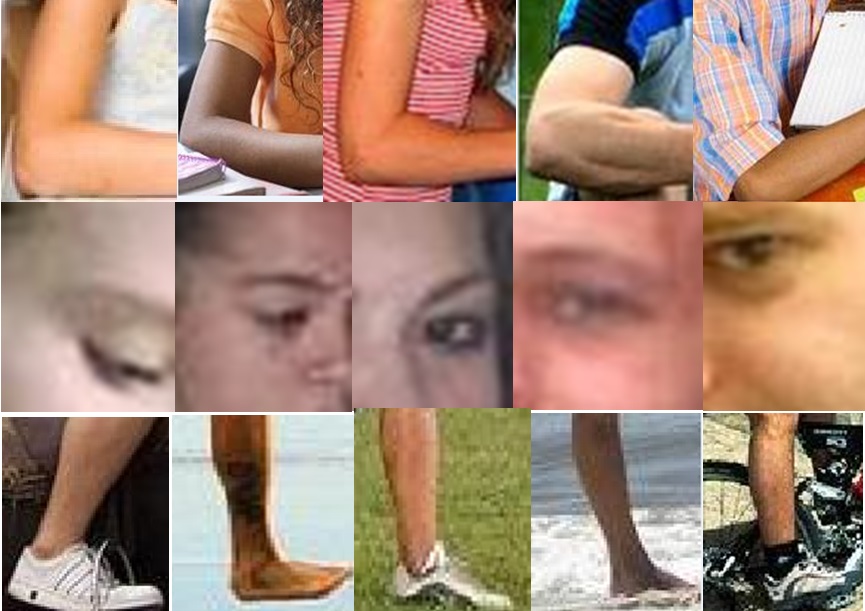}
	\end{center}
	\caption{Three example parts discovered by our fine-tuned CNN. }
	\label{fig:clusterExample}
\end{figure}

After several iterations, we obtain high quality clusters as well as a fine-tuned CNN model which is suitable for our dataset. Each cluster is considered as a part in this paper. Though not all of the clusters can be semantically meaningful, some of them represent particular parts of human body. Fig \ref{fig:clusterExample} shows three example clusters, which are treated as parts in this paper.

We train an SVM classifier for each of the cluster as the part detector. To train the SVM for cluster $ c_i $, we take the patches in this cluster as positive sample set. To be widely representative, the negative sample set not only contains patches from the other clusters but also the patches that are not in any of the clusters.

With these SVM classifiers, an image can be represented by an activation vector of the parts, as well as the spatial locations of these parts. In order to locate parts, we use the sliding window method to scan regions. Even though two images have the same parts, they may come from two different action classes,  such as \textit{fixing a bike} and \textit{riding a bike}. Thus, it is important to model the spatial relationship of parts.

\begin{figure}[htb]
	\begin{center}
		\includegraphics[width=0.8\linewidth]{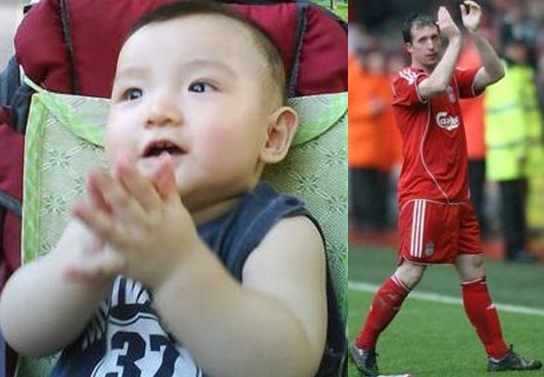}
	\end{center}
	\caption{Two examples from the action class of \textit{applauding} \cite{Yao11humanaction__Stanford40}. The actors are a baby sitting in the baby chair and a man standing in a football court. The appearances of these two actors are quite different.}
	\label{fig:ApplaudingExample}
\end{figure}

If an object has a stable and simple structure, we can model it accurately using the DPM model \cite{DPM_Felzenszwalb:2010:ODD:1850486.1850574} or the constellation model \cite{CONStellation_model}. However, humans involved in the same action class can be quite different due to pose variations. In addition, the appearance of actors varies significantly. For example, in the action class of \textit{applauding}, the actor can be a baby sitting in the baby chair, or a football player standing in the football court, as shown in Fig \ref{fig:ApplaudingExample}. Such variations make the problem of action recognition difficult for traditional part-modeling methods \cite{CONStellation_model,DPM_Felzenszwalb:2010:ODD:1850486.1850574}.

We consider an action as a configuration of parts. Images from the same action class should possess a flexible number of shared spatial configurations of parts. We model the configuration of parts using a newly proposed spatial SPN method introduced in the next subsection.

\begin{figure*}[htb]
	\begin{center}
		\includegraphics[width=0.8\linewidth]{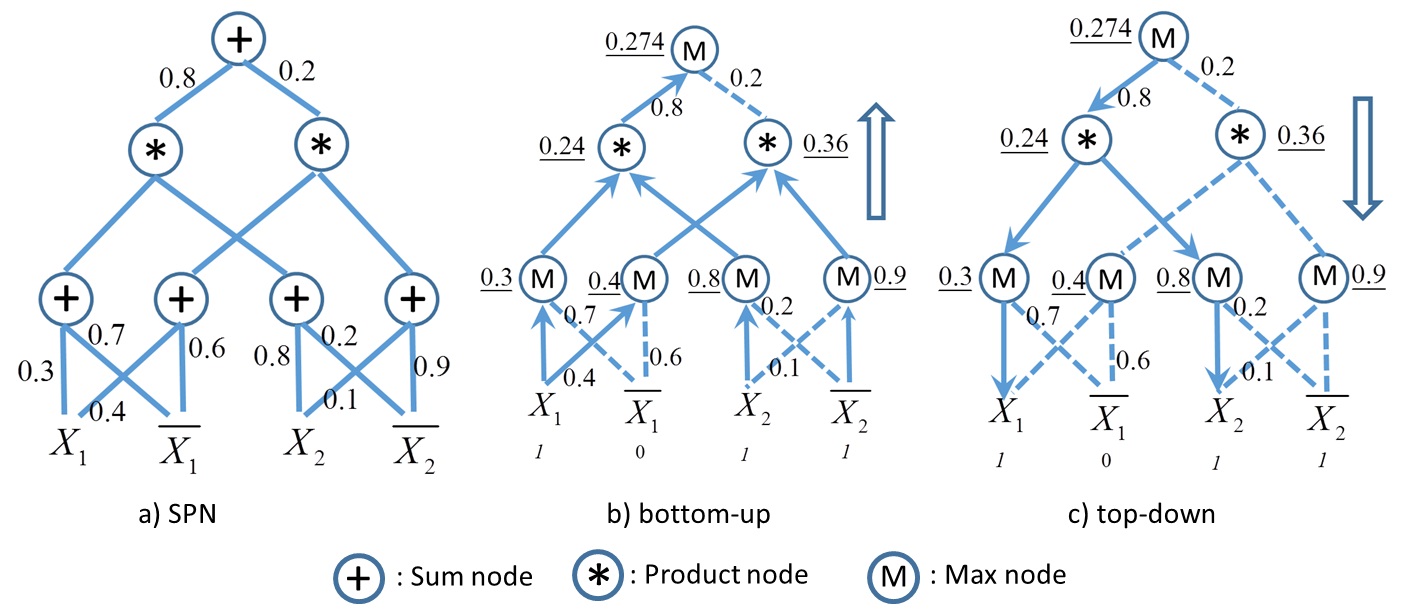}
	\end{center}
	\caption{Examples of SPN. a) shows an SPN with two variables $ X_1 $ and $ X_2 $. b) shows the first two steps to infer the value of $ x_2 $, i.e. bottom-up evaluation with $ X_1=1 $ and marginalized $ X_2 $ (i.e. $ X_2=1 $ and $ \bar{X_2}=1 $) as well as M-node generation. c) shows the third step, i.e. the top-down inference procedure to the value of $ X_2=1 $. }
	\label{fig:SPNExample}
\end{figure*}

\subsection{Hierarchical Spatial SPN}

\subsubsection{Sum Product Networks}

Poon and Domingos \cite{poon2011SPNIntroduce} introduced SPN as a new deep architecture to represent probability distributions  based on the theory of Darwiche's network polynomial \cite{Darwiche:2003:DAI:765568.765570}. 

\textbf{Definition} \cite{poon2011SPNIntroduce} A sum-product network over variables $ x_1,x_2,...,x_d $ is a rooted directed acyclic graph whose leaves are the indicators $ x_1,x_2,...,x_d $ and $ \bar{x}_1,\bar{x}_2,...,\bar{x}_d $ and whose internal nodes are sums and products. Each edge $ (i,j) $ emanating from a sum node $ i $ has a non-negative weight $ w_{ij} $.

SPN is a compact graphical model that allows fast inference and margin computation. We can consider SPN as directed acyclic graphs whose leaves are variables, internal nodes are sums and products \cite{poon2011SPNIntroduce}. These nodes are linked with weighted edges. 
The value of a product node is the product of the values of its children. The value of a sum node is $ \Sigma_{j\in Ch(i)}w_{ij}v_j $, where $ Ch(i) $ are the children of $ i $ and $ v_j $ is the value of node $ j $. The value of an SPN is the value of its root.

In a typical SPN, the parent of a sum node is a product node and the parent of a product node is a sum node \cite{SPNWangXiaogangFacial,poon2011SPNIntroduce}. Fig \ref{fig:SPNExample} a) shows an example of SPN $ S(x_1,\bar{x}_1,x_2,\bar{x}_2) $ over variables $ x_1 $ and $ x_2 $. Based on this SPN, the probability of $ x_1=1 $ and $ x_2=0 $ can be calculated using $ P(x_1,\bar{x}_2)=S(1,0,0,1)=0.8(0.3x_1+0.7\bar{x}_1)(0.8x_2+0.2\bar{x}_2)+0.2(0.4x_1+0.6\bar{x}_1)(0.1x_2+0.9\bar{x}_2)=0.8\times0.3\times0.2+0.2\times0.4\times0.9=0.12 $

Based on an SPN, we can infer the value of an observed variable using MPE (Most Probable Explanation) inference \cite{Darwiche:2003:DAI:765568.765570}. For example, knowing $ x_1=1 $, we can infer the value of $ x_2 $, with three steps shown in Fig \ref{fig:SPNExample} b) and c). The first step marginalizes the unknown variable $ x_2 $ by setting both $ x_2=1 $ and $ \bar{x_2}=1 $ and evaluate SPN accordingly. The second step replaces the sum nodes with M (maximization) nodes and selects the maximum child for each M-node (Fig \ref{fig:SPNExample} b). The third step performs a top-down procedure to track the maximum child for each M node and obtains $ x_2=1 $ (Fig \ref{fig:SPNExample} c).

\begin{figure}[h]
	\begin{center}
		\includegraphics[width=0.8\linewidth]{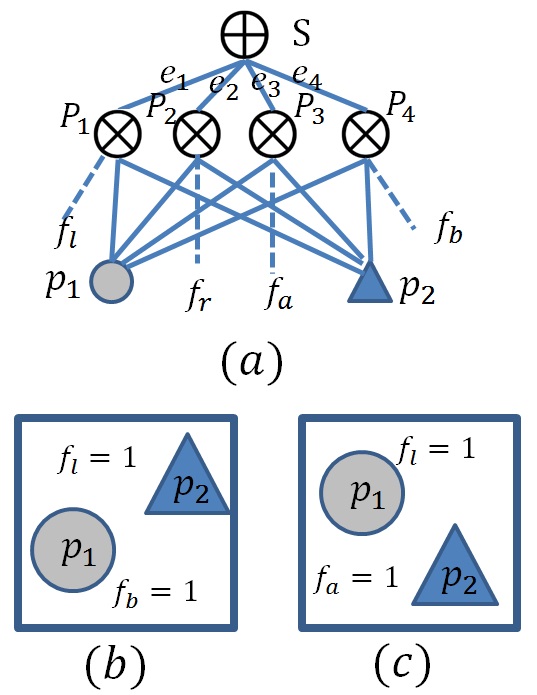}
	\end{center}
	\caption{Indicators $ f_l $, $ f_r $, $ f_a $, and $ f_b $ to capture the spatial relationships of a pair or parts. The indicator $ f_l=1  $ ($ f_r=1 $) means part $ p_1 $ is to the left (right) of part $ p_2 $. The indicator $ f_a=1 $ ($ f_b=1 $) means part $ p_1 $ is above part $ p_2 $. In b), $ f_l=1 $ and $ f_b=1 $; in c) $ f_l=1 $ and $ f_a=1 $. In a) the product nodes $ P_1 $, $ P_2 $, $ P_3 $, and $ P_4 $ capture four different types of spatial relationships of the parts $ p_1 $ and $ p_2 $. The sum node $ S $ combines these four different configurations of part $ p_1 $ and part $ p_2 $ together with the weights of $ e_i (i=1,2,3,4)$. }
	\label{fig:IndicatorforTwoparts}
\end{figure}

\begin{figure}[htb]
	\begin{center}
		\includegraphics[width=0.8\linewidth]{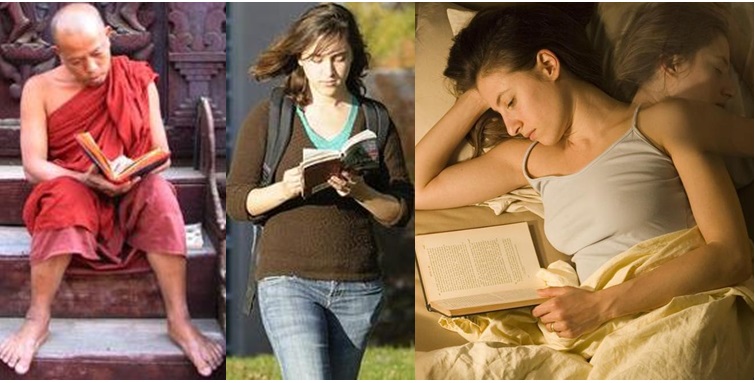}
	\end{center}
	\caption{local spatial relationships between parts are more robust. These three images are from the same action class of \textit{reading}. The actors are in different poses. The long-range spatial relationships between the arms and legs vary a lot.}
	\label{fig:reading_example}
\end{figure}

\begin{figure*}[htb]
	\begin{center}
		\includegraphics[width=0.8\linewidth]{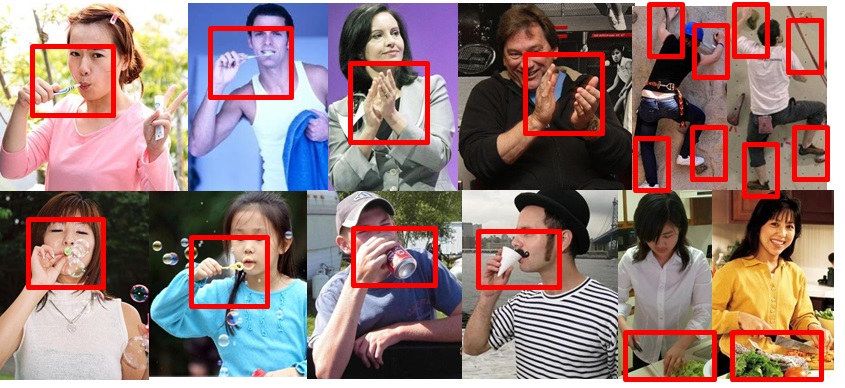}
	\end{center}
	\caption{Samples from six action classes in Stanford40 \cite{Yao11humanaction__Stanford40}: \textit{brushing teeth, applauding, blowing bubbles, climbing, drinking, and cutting vegetables.} The rectangles show informative sub-images of these images.}
	\label{fig:informativePatches}
\end{figure*}

\subsubsection{Hierarchical Spatial Sum Product Networks}

Action recognition in still images is treated as a binary classification problem, where we are given a set of training images $  \{ I_i, i=1,\cdots, N \} $ together with their class labels $ y_i \in \{ 0,1 \} $. Our goal is to learn a spatial SPN for one action class responding more strongly to positive images.

An image is represented by a set of parts (which are learned in \ref{section_part_learning}) $ I_i = (v_i^1,v_i^2, \cdots, v_i^t) \in R^t $, as well as their locations $L_i=(l_i^1,l_i^2, \cdots , l_i^t) \in R^{t\times 2}$, where $ t $ is the number of parts. The binary value $ v_i^j $ indicates whether the $ j $th part occurs in the image $ I_i $ or not. If the $ j $th part occurs ($ v_i^j=1 $), its location is represented by its center pixel $ l_i^j=(x_i^j,y_i^j) $.

The evaluation of conventional SPN proposed in \cite{poon2011SPNIntroduce} is only based on the binary values of its leaf nodes. The important information of spatial relationship between leaf nodes is completely ignored.
To model the spatial relationship between parts, we propose spatial SPN, which is a new SPN structure that can effectively capture the spatial relationships of its leaf nodes. In the proposed spatial SPN, we consider four types of spatial relationship between a pair of parts: \textit{left, right, above, and below}.

In order to model the spatial relationships of two different parts, we introduce an indicator child node for the product node which is the immediate parent of two parts. Fig \ref{fig:IndicatorforTwoparts} shows an example. The indicator child represents the spatial relationships of these two parts. For parts $ p_1 $ and $ p_2 $, we define four indicator variables: $ f_{l} $, $ f_r $, $ f_a $ and $ f_{b} $, respectively denoting part $ p_1 $ is to the left of, to the right of, above and below part $ p_2 $. With these variables, we can capture different types of spatial relationships of these two parts. In Fig \ref{fig:IndicatorforTwoparts} b), part $ p_1 $ is below and left to part $ p_2 $. Thus, the variables $ f_l $ and $ f_b $ equal to $ 1 $; the variables $ f_r $ and $ f_a $ equal to $ 0 $. In Fig \ref{fig:IndicatorforTwoparts} c), part $ p_1 $ is above and to the left of part $ p_2 $. Thus, the variables $ f_a $ and $ f_l $ equal to $ 1 $; the variables $ f_b $ and $ f_r $ equal to $ 0 $. With one of the indicator variables as a child, each of the four product nodes $ P_i(i=1,2,3,4) $ represents a specific spatial relationship between part $ p_1 $ and part $ p_2 $. The sum node $ S $ combines these four spatial configurations together. As we have different nodes to model different spatial relationships, the proposed SPN can deal with deformable spatial relationship between part pairs.
The weight of $ e_i $  represents how likely the $ i $th configuration can be seen in an image of an action class.

With the structure in Fig \ref{fig:IndicatorforTwoparts}, we can model the spatial relationship of every part pairs.
However, it is not preferred to model such pairwise relationship at the whole image scale due to the following two reasons. First, even for the same action class, configuration of parts can vary a lot in images. It is not robust to model long-range pairwise relationship. Fig \ref{fig:reading_example} shows three images from the action class of \textit{reading}. In these three images, the actors are in different poses, i.e. sitting, standing, and lying. The long range spatial relationships between the arms and the legs vary a lot, hence are difficult to model. 
In contrast, local pairwise spatial relationship can be very reliable. We observe that almost all of the action classes have similar configuration of parts in one or more sub-images, as shown in Fig \ref{fig:informativePatches}.  In some cases, we human beings can easily predict the action class of an image based on these informative sub-images. For example, we can recognize the action class of
\textit{cutting vegetables} by a sub-image containing \textit{a knife above vegetables}.

\begin{figure*}[htb]
	\begin{center}
		\includegraphics[width=0.75\linewidth]{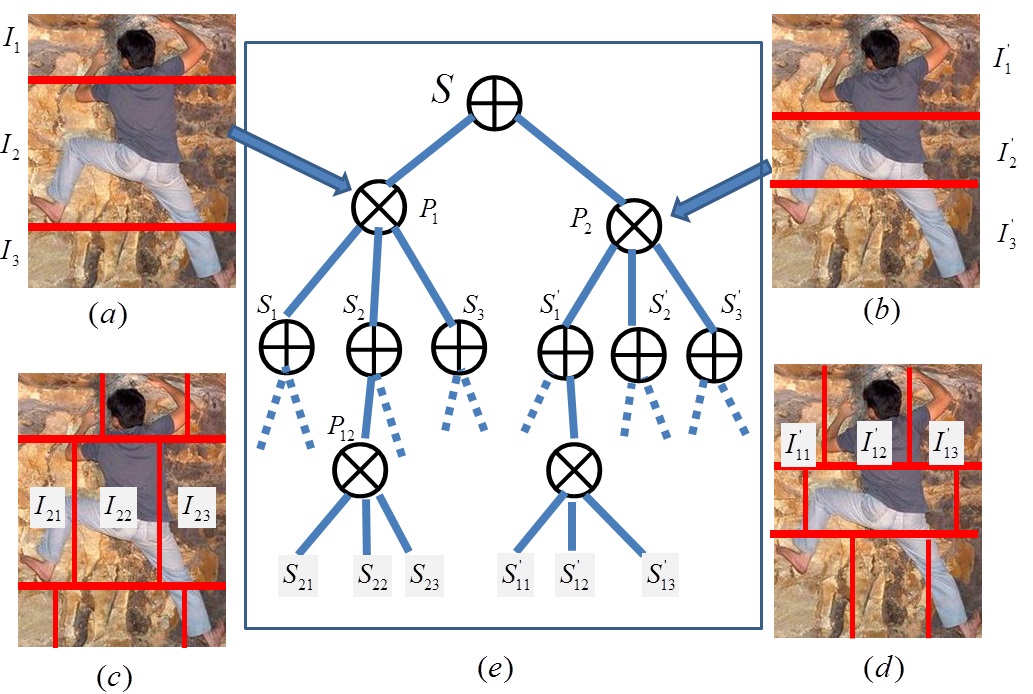}
	\end{center}
	\caption{An example of our hierarchical spatial SPN. Images I is a sample from action class of \textit{climbing} in Stanford40 dataset \cite{Yao11humanaction__Stanford40}. (a) and (b) show two different partitions of the image, corresponding to two product nodes $ P_1 $ and $ P_2 $  in the SPN. The root sum node $ S $ combine the information together which is represented by its children $ P_1 $ and $ P_2 $. Sub-image $ I_2 $ (shown in (c)) is further partitioned into $ I_{21} $, $ I_{22} $, and $ I_{23} $ corresponding to the product node $ P_{12} $.}
	\label{fig:SPNconstructionPartition}
\end{figure*}

Second, it is computationally expensive to model pairwise relationship at the whole image scale.
With $ N $ parts, there are $\slfrac{N(N-1)}{2} $ possible part pairs at the whole image scale.
Such a large number of part pairs will lead to a very large network that is hard to learn. And overfitting may happen. Instead, if we only consider modeling spatial relationships between parts inside a local sub-image, the number of possible part pairs which co-occur in the sub-image can be much smaller.

Based on such intuitions, we aim to model pairwise spatial relationship of parts locally inside sub-images, and drop the long range spatial relationships of part pairs. As shown in Fig \ref{fig:SPNconstructionPartition}, we hierarchically partition an image into sub-images, and model pairwise spatial relationship between parts inside leaf node sub-images only.  We further model the correlations of sub-images using extra SPN layers on the top.

In the proposed hierarchical spatial SPN, we associate a product node with a specific partition and use it to model the correlations of resulting sub-images. (Here, a partition refers to dividing an image into a set of specific regions. For example, an image $ I \in R^{100\times100} $ is partition into three sub-images $ I_1 \in R^{20 \times100} $, $ I_2 \in R^{30 \times100} $, and $ I_3 \in R^{50 \times100} $. In Fig \ref{fig:SPNconstructionPartition}, (a) and (b) are two different partitions, because they generated different sub-images.) We may have a number of different partitions for an image (or sub-image), each resulting in a product node.
The information learned by these partitions are combined together by a sum node, which is the parent of  these product nodes.
In Fig \ref{fig:SPNconstructionPartition}, the product node $ P_1 $ is associated with a partition which divides image $ I $ into $ I_1, I_2 $ and $ I_3 $. This product node $ P_1 $ models the correlations between the sum nodes $ S_1, S_2$ and $ S_3 $ respectively representing these three sub-images $ I_1, I_2 $ and $ I_3 $. The product node $ P_2 $ is associated with another partition method (dividing image $ I $ into $ I_1^{'}, I_2^{'} $ and $ I_3^{'} $).
As the parent of these two product nodes $ P_1 $ and $ P_2 $, the root sum node $ S $ combines the information learned by $ P_1 $ and $ P_2 $. This means the information conveyed by these two different partitions can collaborate in predicting the action class. Similarly, we associate the product node $ P_{21} $ with the partition of dividing $ I_2 $ into $ I_{21} $, $ I_{22} $, and $ I_{23} $. (The sum node $ S_2 $ can have more product children to represent other partition methods of $ I_2 $.) Then, we build an SPN with $ S_{2i} (i=1,2,3)$ as the root node to model the pairwise spatial relationship inside sub-image $ I_{2i} (i=1,2,3)$.

There are a huge number of possible ways of hierarchically partitioning an image into sub-images. We can not model every possible partitions in our spatial SPN. We select a number of discriminant partitions using heuristics before formulating them in spatial SPN. For a specific partition of image $ I $ into $ s $ sub-images, we obtain $ s $ part activation vectors for the sub-images and represent the whole image by the concatenation of these activation vectors. Then, we train a classifier and produce a classification accuracy. Only the partitions with high accuracy scores are considered. As shown in another work \cite{Wang2015Video}, this hierarchical method achieves good performance.

Algorithm \ref{algorithm_spnlearning} shows the procedure to learn the structure of the top three layers of the SPN for each class.

\begin{algorithm}[h]
	\KwData{The part activation vectors $ I_i \in R^t $ for training images, as well as the locations of these parts $L_i \in R^{t\times 2}$, where $ t $ is the number of parts.}
	\KwResult{A spatial SPN　structure for each class }
	\For{each action class}{
		Randomly partition the images into $ s $ sub-images with $ M $ different stretagies\;
		\For {each partition strategy}{
			1. concatenate the BoW representation of each sub-image and obtain an $ s\times t $ representation vector for each image\;
			2. train a classifier and obtain a classification accuracy\;
		}
		Build a product node for each of the partitions corresponding to the $ m (m<M) $ highest classification accuracies\;
	}
	\caption{SPN structure learning algorithm}
	\label{algorithm_spnlearning}
\end{algorithm}

\begin{figure}[htb]
	\begin{center}
		\includegraphics[width=0.95\linewidth]{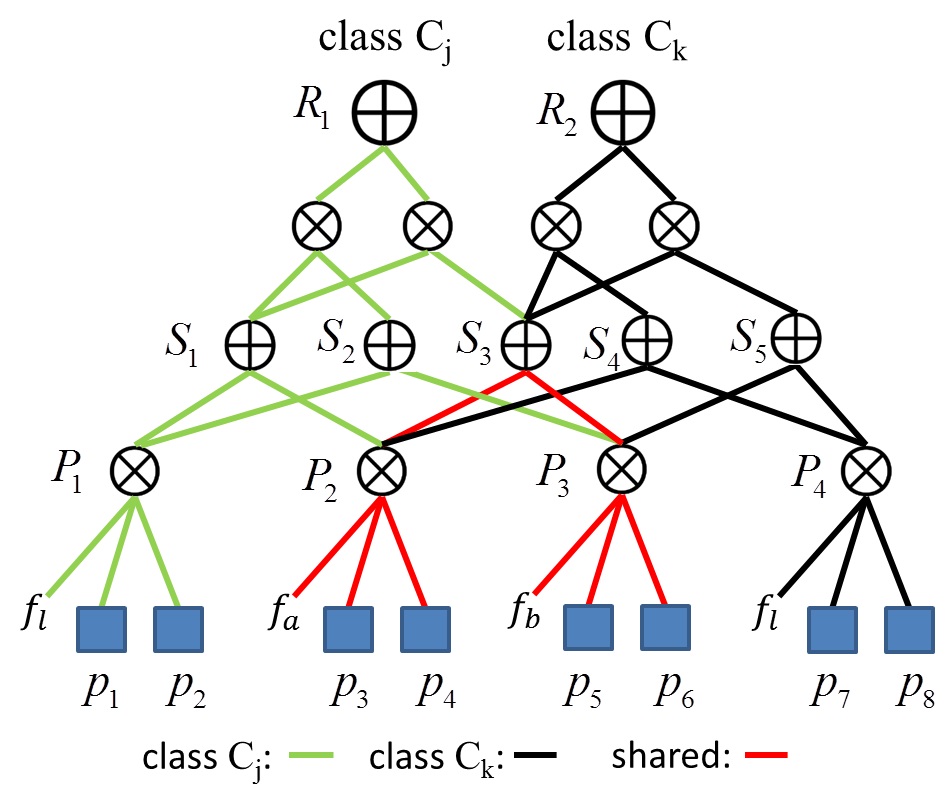}
	\end{center}
	\caption{Shared nodes and edges between two SPNs. A shared sub-image is modeled by the node $ R_1 $ in class $ C_j $ and the node $ R_2 $ in class $ C_k $. The green lines denote the edges for class $ C_j $ and the black lines denote the edges for class $ C_k $. The red lines denote the shared edges for these two classes. They have shared edges because the part pair relationships \textit{part $ p_3 $ above part $ p_4 $} and \textit{part $ p_5 $ below part $ p_6 $} occur in both classes. These two classes have shared product nodes ($ P_2 $ and $ P_3 $) and sum node ($ S_3 $).}
	\label{fig:sharing_in_subimage}
\end{figure}

Normally, algorithm \ref{algorithm_spnlearning} learns different sets of partition methods for different action classes. However, two action classes $ C_j $ and $ C_k $ may share a number of partition methods and have some shared sub-image structures. This means both the SPN for $ C_j $ and the SPN for $ C_k $ need to model the part correlations inside the shared sub-images. If a pair of parts with the same spatial relationship  co-occur inside a shared sub-image structure, the SPNs for two different classes can have shared nodes, as shown in Fig \ref{fig:sharing_in_subimage}. 
In this figure, the root nodes $ R_1 $ and $ R_2 $ respectively represent the shared sub-image structure for class $ C_j $ and class $ C_k $. The leaf nodes represent the parts.
As the part pair relationship \textit{part $ p_3 $ above part $ p_4 $} and \textit{part $ p_5 $ below part $ p_6 $} occur in both classes, the two SPNs have shared edges (denoted by red lines), shared product nodes (i.e. $ P_2 $ and $ P_3 $), and shared sum node (i.e. $ S_3 $). We identify the shared edges and learn their weights based on the images from these two classes.

\subsubsection{Learning}

Our hierarchical spatial SPN takes the part activation vector of an image as well as the locations of these parts as input. 
Let $ S(I_m) $ denote the evaluation of a spatial SPN with the representation of image $ I_m $ as the input. Let $ V(I_m) $ denote the root value of the spatial SPN $ S(I_m) $.
The value $ V(I_m) $ represents the classification score.

After fixing the structure of spatial SPN, we learn the parameters for its edges by MPE (most probable explanation) inference \cite{Discriminative_Learning_SPNNIPS2012_4516}. For an SPN of a specific action class $ C_k $, our objective is to let the value of the root node to be larger for the positive images and smaller for the negative images. To learn the spatial SPN for class $ C_k $, we obtain the following objective function :

\begin{equation}\label{Eq:ClassSPNobjectFunction} 
\begin{split}
min \qquad  &\underset{I_m \in C_k, I_n \notin C_k}{\sum} \xi_{mn}^2 \\
 s.t.\qquad &V(I_m) \geq V(I_n)+1-\xi_{mn}
\end{split}
\end{equation}
where $ \xi_{mn} $ is a slack variable, $ I_m $ is a positive image, and $ I_n $ is a negative image.

Our training procedure has two stages. The first stage improves the representative ability of a spatial SPN. The second stage aims to enhance the discriminative ability of the SPN learned in the first stage.

In the first stage, we train the spatial SPN using a generative algorithm based on inference \cite{poon2011SPNIntroduce}. Algorithm \ref{algorithm_generative_spn} shows the details. After obtaining the SPN using this algorithm, we investigate the weight associating with each edge and delete the edges whose weights equal to zero.

\begin{algorithm}[h]
  \KwData{The part activation vectors, as well as the locations of the parts (if occur), of the images from action class $ C_k $.}
  	\KwResult{A spatial SPN $ S_k $ for action class $ C_k $.}

    \Repeat{convergence or early stopping condition}{
      \For {$ I_m \in C_k $}{
      UpdateWeights($S_k$, Inference($S_k$,$I_m$))
      }
    }
	\caption{SPN parameter learning algorithm}
	\label{algorithm_generative_spn}
\end{algorithm}

In the second stage, we improve the discriminative ability of the learned SPN from the first stage. To achieve this, we take a pair of images from two different classes as input and update the weights of the spatial SPNs for these two classes.

Assume $ S_j $ and $ S_k $ are two different SPNs, with shared tree structure (e.g. red lines in Fig \ref{fig:sharing_in_subimage}), respectively for action classes of $ C_j $ and $ C_k $. Let $ I_m $ and $ I_n $ be two images respectively from the action classes of $ C_j $ and $ C_k $. To update the parameter of SPN $ S_k $, we first evaluate this SPN with the part activation vectors as well as their locations from both $ I_m $ and $ I_n $. Then, in order to overcome gradient diffusion, we convert the two evaluations of SPN to MPN (max-product network), i.e. replacing the sum nodes with M (maximization) nodes, as shown in Fig \ref{fig:SPNExample} from a) to b).

\begin{figure}[htb]
	\begin{center}
		\includegraphics[width=0.95\linewidth]{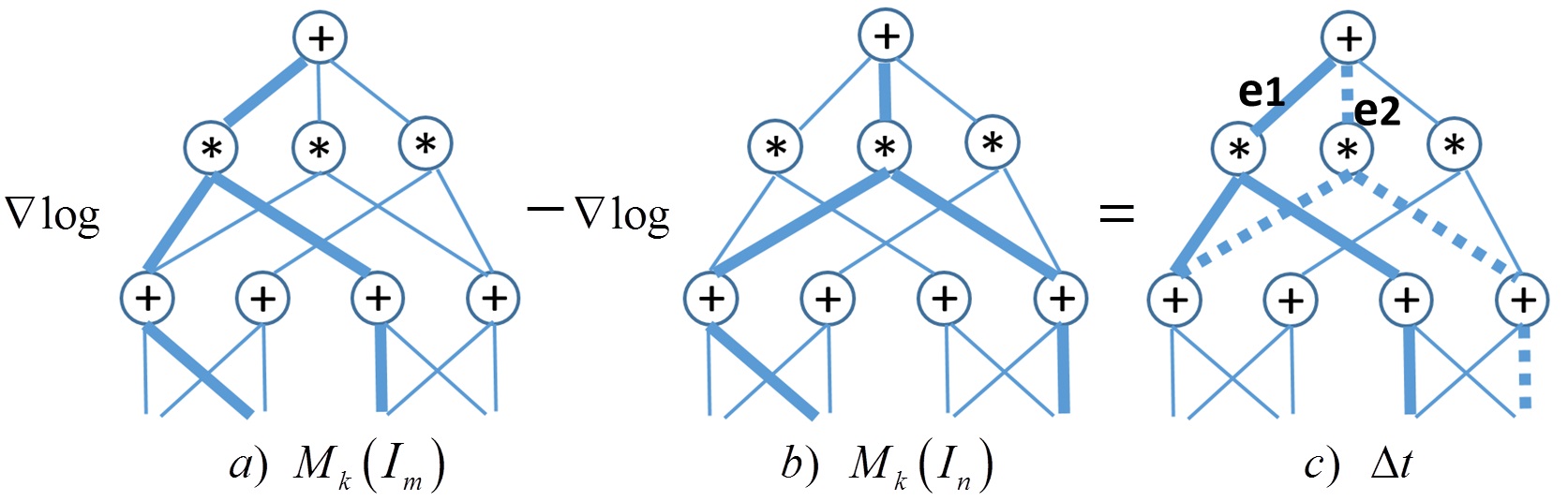}
	\end{center}
	\caption{An example to calculate the partial derivative of the logarithm with respect to the edge weight. a) and b) are respectively the MPNs obtained based on the evaluations on image $ I_m $ and $ I_n $. c) shows the difference between the times that an edge is traversed.}
	\label{fig:differenceSPN}
\end{figure}

Using $ M_k(I_m) $ and $ M_k(I_n) $ to represent these two MPNs (obtained based on the evaluations on $ I_m $ and $ I_n $), the partial derivative of the logarithm with respect to the edge weight $ w_i $ can be calculated as follows
\begin{equation}\label{derivateOfWeight}
\frac{\partial logM_k}{\partial {w_i}}=\frac{\partial logM_k(I_m)}{\partial {w_i}}-\frac{\partial logM_k(I_n)}{\partial {w_i}}=\dfrac{t_i^m}{w_i}-\dfrac{t_i^n}{w_i}
\end{equation}                 
where $ t_i^m $ and  $ t_i^n $ count the times that the $ i $th edge is traversed by the MPE inference path in MPN $ M_k(I_m) $ and MPN $ M_k(I_n) $. The gradient of the log likelihood of the weight is $ \Delta t_i/w_i $, where $ \Delta t_i=t_i^m-t_i^n $ is the difference between the number of times that $ w_i $ is traversed when evaluated on the two images. Fig \ref{fig:differenceSPN} shows an example of this procedure. In this fig, a) and b) respectively show the edges that are traversed in the MPN obtained based on image $ I_m $ and $ I_n $. Fig \ref{fig:differenceSPN} c) shows the partial derivative of the MPN.

In this way, we learn the parameters of the shared tree-structure (e.g. red lines in Fig \ref{fig:sharing_in_subimage}) not only using the images from action class $ C_k $ but also images from $ C_j $. The weights of the edges specifically for class $ C_k $ (black lines in Fig \ref{fig:sharing_in_subimage})   are learned to only favor images from this class to enhance discriminant ability.

\section{Experiments}
\label{Sec:experi}

\subsection{Datasets}

We test the proposed hierarchical spatial SPN on two publicly available datasets: Willow 7 human actions \cite{Delaitre10recognizinghuman} and Stanford 40 human actions \cite{Yao11humanaction__Stanford40}.

Willow 7 human actions \cite{Delaitre10recognizinghuman} is a dataset for image-based action classification. It contains $ 1,791 $ unconstrained consumer images downloaded from the Internet, belonging to $ 7 $ classes of common human actions: \textit{interacting with computer, photographing, playing music, riding bike, riding horse, running, and walking}. Each class has at least $ 108 $ images in total and at least $ 70 $ images for training.

Stanford 40 \cite{Yao11humanaction__Stanford40} is a larger database containing $ 40 $ different types of daily human actions. It has $ 9,352 $ images in total. The number of images for each class ranges from $ 180 $ to $ 300 $. The dataset provides the train and test split for each class, which uses $ 100 $ images of each class for training and the rest for testing.

\begin{table*}[htb]
    \caption {Precision (\%) for each class on the Willow 7 action dataset}   
        
    \begin{center}
        
        \begin{tabular}{c c c c c c c c}
            \hline                        
            \hline
             & Inter.\cite{Delaitre11learningperson-object} &  SP\cite{Lazebnik_beyond_cvpr2006} & ov.SP\cite{Lazebnik_beyond_cvpr2006}  & Dsal.\cite{sharma:CVPR2012_Discriminative}  & FS-SPN & IHS-SPN & JHS-SPN \\ \hline
               InterWComp & 56.6 & 49.4 & 57.8 & 59.7 & 59.3 & 64.2 & 64.2\\
            Photographing & 37.5 &  41.3 & 39.3 & 42.6 & 43.9 & 49.4 & 49.4\\
            playingMusic & 72.0 & 74.3 & 73.8 & 74.6 & 72.4 & 76.2 & 76.2\\
            RidingBike & 90.0 & 87.8 & 88.4 & 87.8 & 86.3 & 94.6 & 95.2\\
            RidingHorse & 75.0 & 73.6 & 80.8 & 84.2 & 81.3 & 85.1 & 85.6\\
            Running & 59.7 &  53.3 & 55.8 & 56.1 & 57.6 & 65.4 & 66.0\\
            Walking & 57.6 & 58.3 & 56.3 & 56.5 & 56.9 & 64.5 & 65.1\\
            \hline
                        
                        \hline
       \end{tabular}
        
    \end{center}

    \label{tab:class_willow_accuracy} 
\end{table*}

\subsection{Part Discovery}

We discover parts by iteratively conducting three steps: feature extraction, clustering, and fine-tuning. In the first step, we represent the patches using deep features extracted from the CNN model. In the second step, we obtain a tentative label for a patch by unsupervised clustering. In the third step, we fine tune the CNN model to fit our data, and obtain better representations of the patches in the next iteration.

Firstly, to discover the parts from the images, we densely sample $ 1,000 $ patches from each training image of Stanford 40 dataset. 
We resize these patches to be the same size and take them as the inputs of the pre-trianed CNN model on imageNet \cite{jia2014caffe}. We extract the $ 4,096 $ dimensional features of the first fully-connected layer.

Then, we perform K-means clustering on the deep features of all patches and obtain a set of over-segmented clusters.  These clusters are agglomerated into $ N_c $ centers based on average link \cite{refine_cluster} to capture the spherical structure. The average link between two clusters \textbf{$ C_1 $} and \textbf{$ C_2 $} are calculated as \begin{equation}\label{Cluster_likn} D(C_1,C_2)=\frac{1}{|C_1|.|C_2|}\underset{x\in C_1}{\sum }\underset{y\in C_2}{\sum }d(x,y) \end{equation}  where $ d(x,y) $ measures the distance between $ x $ and $ y $. The closest two over-segmented clusters are merged together until the number of centers reduces to $ N_c $. In this procedure, we drop the clusters which are small and far from the rest. Each of these $ N_c $ centers corresponds to a tentative data-driven attribute. The parameter $ N_c $ gradually reduces from $ 2,000 $ to $ 500 $ in our experiments.

Thirdly, with the tentative cluster labels and the data, we adapt the CNN model to this visual pattern discovery task via fine-tunning. In the fine-tunned model, the soft-max layer has  $ N_c $ nodes, each corresponding to a cluster. We treat patches from one cluster as the positive training data of the corresponding node. In this way, we obtain an adapted CNN model for these clusters without supervision. The adapted CNN model is expected to generate more suitable feature representation for our task.

After obtaining the clusters by $ 10 $ iterations, we train a SVM classifier for each cluster. In the testing stage, we densely sample patches with a fixed step size of $ 4 $ pixels to increase the variety of patches. As the proposed method hierarchically partition the image into sub-images and model the spatial relationships of these parts inside the sub-images, we restrict that one part occurs no more than one time in a sub-image. The part location is represented by the coordinate of the center point of the patch that produces the highest detection score.

\subsection{Action Classification}

Based on the edge weights of the learned SPN, we observe that all of the sub-images are equally representative in the action classes of \textit{jumping} and \textit{climbing}. However, in some action classes, some sub-images are much more representative than the others, such as the bottom sub-images in the action class of \textit{cleaning the floor} and the top sub-images in the action class of \textit{drinking}.

\begin{table}[h!]
    \caption {Mean average precision on Willow 7 action dataset}   
    \begin{center}
        
        \begin{tabular}{c c c c c}
            \hline
            
            \hline
             Inter. & Dsal & SPM  & EPM  & EPM+context \\
            \cite{Delaitre11learningperson-object} & \cite{sharma:CVPR2012_Discriminative} & \cite{Lazebnik_beyond_cvpr2006} & \cite{sharma:CVPR2013} & \cite{sharma:CVPR2013}\\
        \hline
     64.1\% & 65.9\% & 63.7\% & 66.0\% & 67.6\%\\
			\hline
			\hline
			
			\hline
SPN & FS-SPN & IHS-SPN & JHS-SPN \\ \hline
48.7\% & 65.3\% & 71.3\% & 71.7\%\\ \hline  

 \hline
       \end{tabular}
        
    \end{center}

    \label{tab:willow_recognition_accuracy} 
\end{table}

\begin{table}[h!]
		\caption {Mean average precision on Stanford 40 action dataset} 
	\begin{center}
		
		\begin{tabular}{c c c c c}
			\hline
			
			\hline
			Object & LLC & SPM  & EPM  & EPM+context \\
			bank \cite{obect_bank} & \cite{LLC} & \cite{Lazebnik_beyond_cvpr2006} & \cite{sharma:CVPR2013} & \cite{sharma:CVPR2013}\\
			\hline
			
			32.5\% & 35.2\% & 34.9\% & 40.7\% & 42.2\% \\ 
			\hline
			\hline
			
			\hline
			SPN & FS-SPN & IHS-SPN & JHS-SPN & \\ \hline 
			32.8\% & 41.5\% & 43.1\% & 44.3\% &  \\ \hline 
			
			\hline
			
		\end{tabular}
		
	\end{center}

	\label{tab:stanford_recognition_accuracy} 
\end{table}

To show the effectiveness of the proposed hierarchical spatial SPN (HS-SPN), we compare it with other two different SPN structures. The first one is the naive SPN that does not take the spatial information into consideration. The second one is the flat spatial SPN (FS-SPN) structure that considers not only local pairwise spatial relationships, but also long-range pairwise spatial relationships without the hierarchical partition method. For the HS-SPN, we have two different learning strategies: individual learning HS-SPN (IHS-SPN) and joint learning HS-SPN (JHS-SPN). The IHS-SPN does not consider the shared sub-images and the shared edges between SPNs for two different classes. The JHS-SPN learns the weights of the shared edges (e.g. red lines in Fig \ref{fig:sharing_in_subimage}) jointly using the images from two different classes.

Table \ref{tab:class_willow_accuracy} lists the classification accuracies of different methods on the seven action classes of Willow 7.
Table \ref{tab:willow_recognition_accuracy} and table \ref{tab:stanford_recognition_accuracy} list the mean average precisions of different methods respectively on the Willow 7 and Stanford 40 datasets. Besides our baselines, we also list the accuracies of some recent methods, including Inter. \cite{Delaitre11learningperson-object}, Dsal. \cite{sharma:CVPR2012_Discriminative}, SPM (spatial pyramid method) \cite{Lazebnik_beyond_cvpr2006}, and two EPM (expanded parts model) methods \cite{sharma:CVPR2013}. Our method outperforms the latest method \cite{sharma:CVPR2013} by 4.1\% and 2.1\% respectively on Willow 7 dataset and Stanford 40 dataset.

As we can see from table \ref{tab:willow_recognition_accuracy} and table \ref{tab:stanford_recognition_accuracy}, the FS-SPN significantly outperforms SPN (17.0\% on Willow 7 and 11.3\% on Stanford 40). This means the spatial relationship is indeed important for action recognition. The classification accuracy of naive SPN is only 21.2\% on the action class of \textit{fixing a bike}. Most of the images in this class are misclassified into the class of \textit{riding a bike}. This is because the same set of parts occur in these two action classes (\textit{fixing a bike} and \textit{riding a bike}). Similarly, SPN cannot correctly classify the images from the action classes of \textit{riding a horse} and \textit{feeding a horse}. Incorporating the spatial information, the proposed spatial SPN increases the accuracies of these action classes significantly (from 21.2\% to 76.2\% for \textit{riding a bike} and from 10.7\% to 44.3\% for \textit{riding a horse}).

\begin{figure}[htb]
	\begin{center}
		\includegraphics[width=0.8\linewidth]{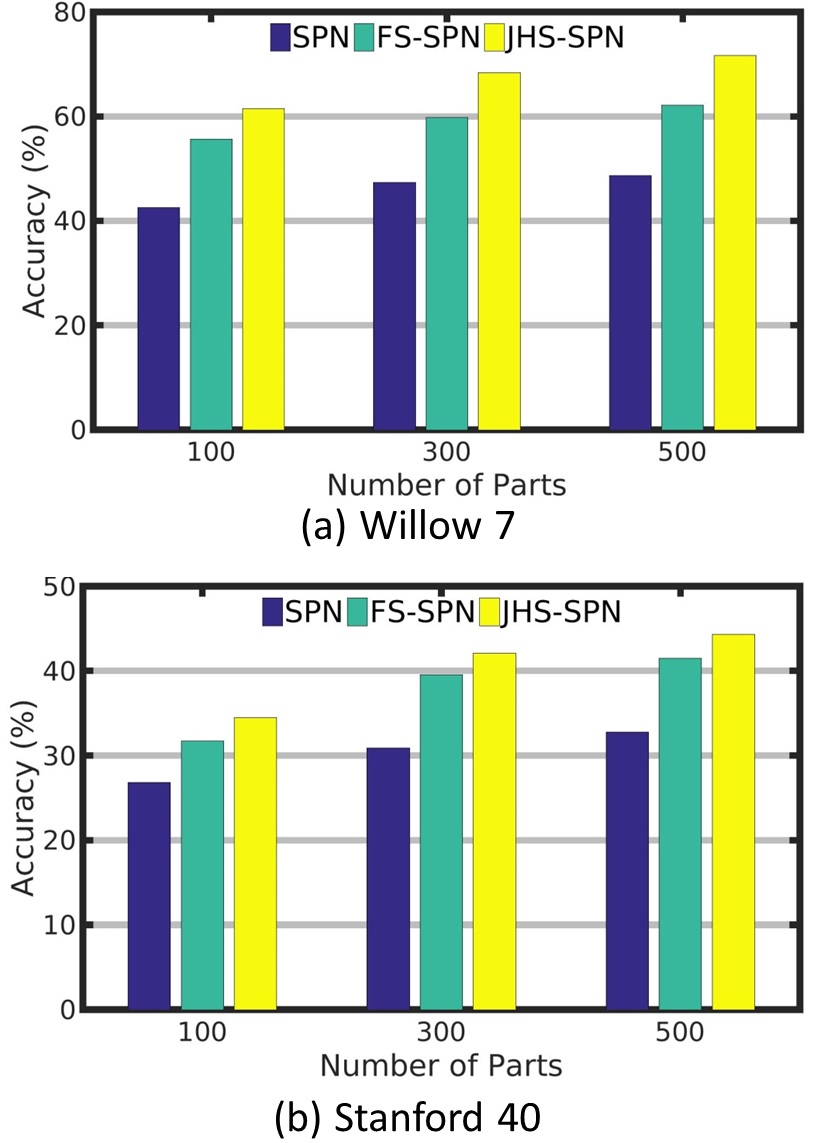}
	\end{center}
	\caption{The action recognition varies with the number of parts on Willow 7 dataset and Stanford 40 dataset. (SPN: the naive SPN proposed in \cite{poon2011SPNIntroduce} without spatial information. FS-SPN: Flat Spatial SPN that considers not only local spatial relationship but also long-range spatial relationship between parts. JHS-SPN: Joint Learning Hierarchical Spatial SPN.)}
	\label{fig:AccNumberofParts}
\end{figure}

The experimental results also show that HS-SPN performs better than FS-SPN. This demonstrates the effectiveness of our hierarchical partition method. With $ 500 $ parts on Willow 7 dataset, while the FS-SPN needs to model $ 249,500 $ part pairs, IHS-SPN and JHS-SPN only need to model $ 5,000 $ part pairs after hierarchical partition. This significantly simplifies the structure of the learned SPN and reduces the computational complexity.

The JHS-SPN achieves accuracy 0.4\% and 1.2\% higher than IHS-SPN on respectively Willow 7 dataset and Stanford 40 dataset. This means the joint learning is useful in this task. The improvement of accuracy mainly originates from the action classes that have shared structure with some other classes. In Willow 7, the action classes that share many nodes in the learned SPN are: \textit{running} and \textit{walking}, \textit{riding a bike} and \textit{riding a horse}. The action classes \textit{running} and \textit{walking} share the top sub-images. This is because the upper body of the actors are quite similar in the images of these two different action classes.

Fig \ref{fig:AccNumberofParts} shows how the action recognition accuracy varies with the number of parts. Normally, we can improve the accuracy by discovering more parts. With $ 100 $, $ 300 $, and $ 500 $ parts, the recognition accuracies of JHS-SPN are respectively 32.8\%, 41.5\%, and 44.3\% on Stanford 40 dataset. While the $ 200 $ parts from 101-300 improve the accuracy by 8.7\%, the parts from 301-500 only improve the accuracy by 2.8\%.

\begin{figure}[htb]
	\begin{center}
		\includegraphics[width=0.95\linewidth]{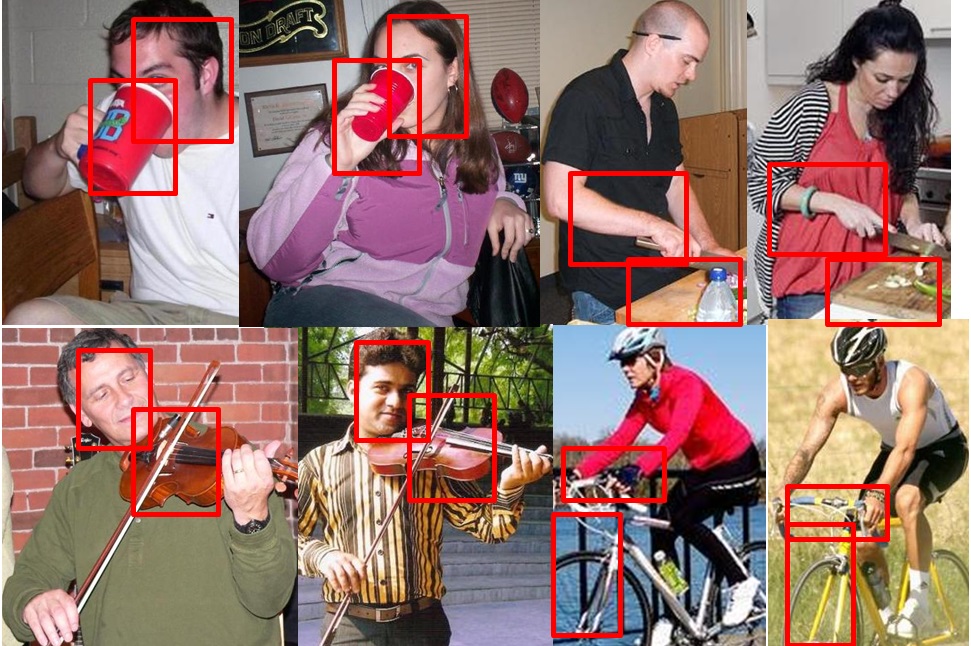}
	\end{center}
	\caption{The discriminative part pairs discovered by our learned hierarchical spatial SPN.}
	\label{fig:finding_pairs}
\end{figure}

We also discover discriminative part pairs based on the learned hierarchical SPN, as shown in Fig \ref{fig:finding_pairs}. To measure the discriminative ability of a part pair, we disable this part pair to see the drop of the accuracy. A larger drop in accuracy indicates a more discriminative part pair.

\section{Conclusion}
\label{Sec:Conclusion}

In this paper, we propose hierarchical spatial SPN for action recognition from still images. In the proposed method, we introduce indicator children for the product nodes to model the spatial relationships of part pairs. Also, we propose to encode the discriminant partitions of images using SPN. The experimental results proves the effective of the proposed method.

\section*{Acknowledgment}
The research is supported by Singapore Ministry of Education (MOE) Tier 2 ARC28/14, and Singapore A*STAR Science and Engineering Research Council PSF1321202099.

This work was carried out at the Rapid-Rich Object Search
(ROSE) Lab at the Nanyang Technological University, Singapore.
The ROSE Lab is supported by a grant from the
Singapore National Research Foundation and administered by
the Interactive \& Digital Media Programme Office at the
Media Development Authority.

\ifCLASSOPTIONcaptionsoff
  \newpage
\fi

\bibliographystyle{IEEEtran}
\bibliography{still_action_bib}

\begin{thebibliography}{10}
\providecommand{\url}[1]{#1}
\csname url@samestyle\endcsname
\providecommand{\newblock}{\relax}
\providecommand{\bibinfo}[2]{#2}
\providecommand{\BIBentrySTDinterwordspacing}{\spaceskip=0pt\relax}
\providecommand{\BIBentryALTinterwordstretchfactor}{4}
\providecommand{\BIBentryALTinterwordspacing}{\spaceskip=\fontdimen2\font plus
\BIBentryALTinterwordstretchfactor\fontdimen3\font minus
  \fontdimen4\font\relax}
\providecommand{\BIBforeignlanguage}[2]{{%
\expandafter\ifx\csname l@#1\endcsname\relax
\typeout{** WARNING: IEEEtran.bst: No hyphenation pattern has been}%
\typeout{** loaded for the language `#1'. Using the pattern for}%
\typeout{** the default language instead.}%
\else
\language=\csname l@#1\endcsname
\fi
#2}}
\providecommand{\BIBdecl}{\relax}
\BIBdecl

\bibitem{computers2020088}
\BIBentryALTinterwordspacing
S.-R. Ke, H.~L.~U. Thuc, Y.-J. Lee, J.-N. Hwang, J.-H. Yoo, and K.-H. Choi, ``A
  review on video-based human activity recognition,'' \emph{Computers}, vol.~2,
  no.~2, p.~88, 2013. [Online]. Available:
  \url{http://www.mdpi.com/2073-431X/2/2/88}
\BIBentrySTDinterwordspacing

\bibitem{TCSCVT_1}
X.~Wu, D.~Xu, L.~Duan, J.~Luo, and Y.~Jia, ``Action recognition using
  multilevel features and latent structural svm,'' \emph{Circuits and Systems
  for Video Technology, IEEE Transactions on}, vol.~23, no.~8, pp. 1422--1431,
  Aug 2013.

\bibitem{TCSVT_2}
A.~Haq, I.~Gondal, and M.~Murshed, ``On temporal order invariance for
  view-invariant action recognition,'' \emph{Circuits and Systems for Video
  Technology, IEEE Transactions on}, vol.~23, no.~2, pp. 203--211, Feb 2013.

\bibitem{TCSVT_3}
Z.~Zhang, C.~Wang, B.~Xiao, W.~Zhou, and S.~Liu, ``Cross-view action
  recognition using contextual maximum margin clustering,'' \emph{Circuits and
  Systems for Video Technology, IEEE Transactions on}, vol.~24, no.~10, pp.
  1663--1668, Oct 2014.

\bibitem{TCSVT_4}
T.~Nguyen, Z.~Song, and S.~Yan, ``Stap: Spatial-temporal attention-aware
  pooling for action recognition,'' \emph{Circuits and Systems for Video
  Technology, IEEE Transactions on}, vol.~25, no.~1, pp. 77--86, Jan 2015.

\bibitem{reviewer1_6918650}
P.~Foggia, A.~Saggese, N.~Strisciuglio, and M.~Vento, ``Exploiting the deep
  learning paradigm for recognizing human actions,'' in \emph{Advanced Video
  and Signal Based Surveillance (AVSS), 2014 11th IEEE International Conference
  on}, 2014, pp. 93--98.

\bibitem{reviewer1_Baccouche:2011:SDL:2177908.2177914}
M.~Baccouche, F.~Mamalet, C.~Wolf, C.~Garcia, and A.~Baskurt, ``Sequential deep
  learning for human action recognition,'' in \emph{Proceedings of the Second
  International Conference on Human Behavior Unterstanding}, ser. HBU'11, 2011,
  pp. 29--39.

\bibitem{reviewer1_Dobhal2015178}
T.~Dobhal, V.~Shitole, G.~Thomas, and G.~Navada, ``Human activity recognition
  using binary motion image and deep learning,'' \emph{Procedia Computer
  Science}, vol.~58, pp. 178 -- 185, 2015.

\bibitem{StillImageSurvey}
G.~Guo and A.~Lai, ``A survey on still image based human action recognition,''
  \emph{Pattern Recognition}, vol.~47, no.~10, pp. 3343 -- 3361, 2014.

\bibitem{YangWMCVPR10}
W.~Yang, Y.~Wang, and G.~Mori, ``Recognizing human actions from still images
  with latent poses,'' in \emph{CVPR}, 2010.

\bibitem{Yao_modeling_mutual_2010_CVPR}
B.~Yao and L.~Fei-Fei, ``Modeling mutual context of object and human pose in
  human-object interaction activities,'' in \emph{CVPR}, 2010.

\bibitem{Delaitre11learningperson-object}
V.~Delaitre, J.~Sivic, and I.~Laptev, ``Learning person-object interactions for
  action recognition in still images,'' 2011.

\bibitem{GroupletYaoBangpengFeifeiLI}
B.~Yao and L.~Fei~Fei, ``Grouplet: A structured image representation for
  recognizing human and object interactions,'' in \emph{CVPR}, 2010.

\bibitem{sharma:CVPR2012_Discriminative}
G.~Sharma, F.~Jurie, and C.~Schmid, ``Discriminative spatial saliency for image
  classification,'' in \emph{CVPR}, 2012.

\bibitem{Delaitre10recognizinghuman}
V.~Delaitre, I.~Laptev, and J.~Sivic, ``Recognizing human actions in still
  images: a study of bag-of-features and part-based representations,'' 2010.

\bibitem{sharma:CVPR2013}
G.~Sharma, F.~Jurie, and C.~Schmid, ``Expanded parts model for human attribute
  and action recognition in still images,'' in \emph{CVPR}, 2013.

\bibitem{Yao11humanaction__Stanford40}
B.~Yao, X.~Jiang, A.~Khosla, A.~L. Lin, L.~J. Guibas, and L.~Fei-Fei, ``Action
  recognition by learning bases of action attributes and parts,'' in
  \emph{ICCV}, 2011.

\bibitem{MajiActionCVPR11}
S.~Maji, L.~Bourdev, and J.~Malik, ``Action recognition from a distributed
  representation of pose and appearance,'' in \emph{CVPR}, 2011.

\bibitem{BourdevMalikICCV09}
L.~Bourdev and J.~Malik, ``Poselets: Body part detectors trained using 3d human
  pose annotations,'' in \emph{ICCV}, 2009.

\bibitem{desai10_action}
C.~Desai, D.~Ramanan, and C.~Fowlkes, ``{Discriminative models for static
  human-object interactions},'' in \emph{CVPR workshop}, 2010.

\bibitem{poon2011SPNIntroduce}
H.~Poon and P.~Domingos, ``Sum-product networks: A new deep architecture,'' in
  \emph{ICCV Workshops}, 2011.

\bibitem{ikizlercinbisICCV2009}
N.~Ikizler-Cinbis, R.~G. Cinbis, and S.~Sclaroff, ``Learning actions from the
  web,'' in \emph{ICCV}, 2009.

\bibitem{Thurau-HlavacPosePrimitivesCVPR2008}
C.~Thurau and V.~Hlavac, ``Pose primitive based human action recognition in
  videos or still images,'' in \emph{CVPR}, 2008.

\bibitem{DBLP:conf/cvpr/WangJDLM06}
Y.~Wang, H.~Jiang, M.~S. Drew, Z.~Li, and G.~Mori, ``Unsupervised discovery of
  action classes,'' in \emph{CVPR}, 2006.

\bibitem{FelzenszwalbMR_CVPR_2008}
P.~Felzenszwalb, D.~McAllester, and D.~Ramanan, ``A discriminatively trained,
  multiscale, deformable part model,'' in \emph{CVPR}, 2008.

\bibitem{BourdevPoseletsECCV10}
L.~Bourdev, S.~Maji, T.~Brox, and J.~Malik, ``Detecting people using mutually
  consistent poselet activations,'' in \emph{ECCV}, 2010.

\bibitem{Delalleau11shallowvs}
O.~Delalleau and Y.~Bengio, ``Shallow vs. deep sum-product networks,'' 2011.

\bibitem{DirectIndirectSPNicml2014c1_rooshenas14}
A.~Rooshenas and D.~Lowd, ``Learning sum-product networks with direct and
  indirect variable interactions,'' in \emph{ICML-14}, 2014.

\bibitem{Discriminative_Learning_SPNNIPS2012_4516}
R.~Gens and P.~Domingos, ``Discriminative learning of sum-product networks,''
  in \emph{NIPS}, 2012.

\bibitem{SPNWangXiaogangFacial}
P.~Luo, X.~Wang, and X.~Tang, ``A deep sum-product architecture for robust
  facial attributes analysis,'' in \emph{ICCV}, 2013.

\bibitem{SPN_video_action}
M.~R. Amer and S.~Todorovic, ``Sum-product networks for modeling activities
  with stochastic structure,'' in \emph{CVPR}, 2012.

\bibitem{liuting}
T.~Liu, G.~Wang, and Q.~Yang, ``Real-time part-based visual tracking via
  adaptive correlation filters,'' in \emph{2015 IEEE Conference on Computer
  Vision and Pattern Recognition (CVPR)}, 2015, pp. 4902--4912.

\bibitem{YiYangCVPR2011:APE:2191740.2192012}
Y.~Yang and D.~Ramanan, ``Articulated pose estimation with flexible
  mixtures-of-parts,'' in \emph{CVPR}, 2011.

\bibitem{Shahroudy2015Multimodal}
A.~Shahroudy, T.~T. Ng, Q.~Yang, and G.~Wang, ``Multimodal multipart learning
  for action recognition in depth videos,'' \emph{IEEE Transactions on Pattern
  Analysis and Machine Intelligence}, vol.~PP, no.~99, pp. 1--1, 2015.

\bibitem{Zuo2015Learning}
Z.~Zuo, B.~Shuai, G.~Wang, X.~Liu, X.~Wang, and B.~Wang, ``Learning contextual
  dependencies with convolutional hierarchical recurrent neural networks,''
  \emph{Computer Science}, 2015.

\bibitem{Shuai2016Scene}
B.~Shuai, Z.~Zuo, G.~Wang, and B.~Wang, ``Scene parsing with integration of
  parametric and non-parametric models.'' \emph{IEEE Transactions on Image
  Processing}, vol.~25, no.~5, pp. 1--1, 2016.

\bibitem{CNN__NIPS2012_4824}
A.~Krizhevsky, I.~Sutskever, and G.~E. Hinton, ``Imagenet classification with
  deep convolutional neural networks,'' in \emph{NIPS}, 2012.

\bibitem{jia2014caffe}
Y.~Jia, E.~Shelhamer, J.~Donahue, S.~Karayev, J.~Long, R.~Girshick,
  S.~Guadarrama, and T.~Darrell, ``Caffe: Convolutional architecture for fast
  feature embedding,'' \emph{arXiv preprint arXiv:1408.5093}, 2014.

\bibitem{DPM_Felzenszwalb:2010:ODD:1850486.1850574}
P.~F. Felzenszwalb, R.~B. Girshick, D.~McAllester, and D.~Ramanan, ``Object
  detection with discriminatively trained part-based models,'' \emph{TPAMI},
  vol.~32, no.~9, 2010.

\bibitem{CONStellation_model}
R.~Fergus, P.~Perona, and A.~Zisserman, ``Weakly supervised scale-invariant
  learning of models for visual recognition,'' \emph{IJCV}, vol.~71, no.~3, pp.
  273--303, 2007.

\bibitem{Darwiche:2003:DAI:765568.765570}
A.~Darwiche, ``A differential approach to inference in bayesian networks,''
  \emph{J. ACM}, vol.~50, no.~3, pp. 280--305, May 2003.

\bibitem{Wang2015Video}
L.~Wang, T.~Liu, G.~Wang, K.~L. Chan, and Q.~Yang, ``Video tracking using
  learned hierarchical features.'' \emph{IEEE Transactions on Image Processing
  A Publication of the IEEE Signal Processing Society}, vol.~24, no.~4, pp.
  1424--35, 2015.

\bibitem{Lazebnik_beyond_cvpr2006}
S.~Lazebnik, C.~Schmid, and J.~Ponce, ``Beyond bags of features: Spatial
  pyramid matching for recognizing natural scene categories,'' in \emph{CVPR},
  2006.

\bibitem{refine_cluster}
R.~R. Sokal and C.~D. Michener, ``{A statistical method for evaluating
  systematic relationships},'' in \emph{University of Kansas Scientific
  Bulletin}, 1958.

\bibitem{obect_bank}
L.-J. Li, H.~Su, E.~P. Xing, and F.-F. Li, ``Object bank: A high-level image
  representation for scene classification semantic feature sparsification.'' in
  \emph{NIPS}, 2010.

\bibitem{LLC}
J.~Wang, J.~Yang, K.~Yu, F.~Lv, T.~Huang, and Y.~Gong, ``Locality-constrained
  linear coding for image classification,'' in \emph{CVPR}, 2010.

\end{thebibliography}

\begin{IEEEbiography}[{\includegraphics[width=1in,height=1.25in,clip,keepaspectratio]{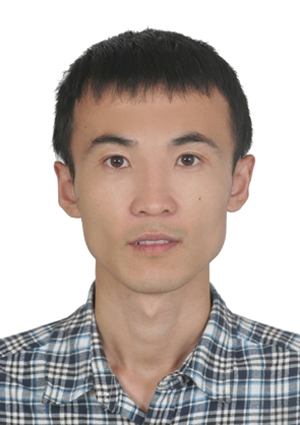}}]{Jinghua Wang}
received his B.Eng. degree from Shandong University, M.S. degree from the Harbin Institute of Technology, and the Ph.D. degree from The Hong Kong Polytechnic University. His a research fellow in Nanyang Technological University. His research interests
include computer vision and
machine learning.

\end{IEEEbiography}

\begin{IEEEbiography}[{\includegraphics[width=1in,height=1.25in,clip,keepaspectratio]{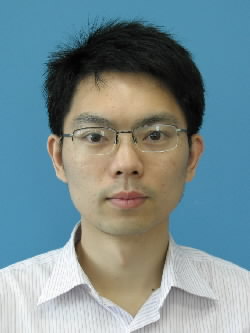}}]{Gang Wang}
is an Assistant Professor with the
School of Electrical and Electronic Engineering at
Nanyang Technological University (NTU), and a
research scientist at the Advanced Digital Science
Center. He received his B.S. degree from Harbin
Institute of Technology in Electrical Engineering in
2005 and the PhD degree in Electrical and Computer
Engineering, University of Illinois at UrbanaChampaign
in 2010. During his PhD study, he is a recipient of the prestigious Harriett \& Robert Perry
Fellowship (2009-2010) and CS/AI award (2009) at
UIUC. His research interests include computer vision and machine learning.
Particularly, he is focusing on object recognition, scene analysis, large scale
machine learning, and deep learning. He is a member of IEEE.
\end{IEEEbiography}

\end{document}